\RequirePackage{fix-cm}
\documentclass[smallextended]{svjour3}       % onecolumn (second format)
\smartqed  % flush right qed marks, e.g. at end of proof
\usepackage{graphicx}
\usepackage{geometry}
\usepackage{pdflscape}
\usepackage{amsmath,amssymb}
\DeclareMathOperator{\E}{\mathbb{E}}
\usepackage{bbm}
\usepackage{enumitem}
\usepackage{fixltx2e}
\usepackage{mathtools}
\usepackage{multirow}
\usepackage{ctable}
\usepackage{mathrsfs}
\usepackage[linesnumbered,ruled]{algorithm2e}% http://ctan.org/pkg/algorithms
\SetEndCharOfAlgoLine{}
\SetKwComment{Comment}{$\triangleright$\ }{}
\usepackage{algpseudocode}% http://ctan.org/pkg/algorithmicx
\usepackage{balance}
\usepackage{fixltx2e}
\usepackage{mathtools}
\usepackage{ctable}
\usepackage{enumitem}
\usepackage[authoryear]{natbib}

\begin{document}
\title{Beyond Word Embeddings: Learning Entity and Concept Representations from Large Scale Knowledge Bases\thanks{In this paper, we use the terms "concept" and "entity" interchangeably.}}


%\titleheader{Accepted at 2017 IEEE International Conference on Big Data (BIGDATA)}

\author{Walid Shalaby         \and
        Wlodek Zadrozny \and 
        Hongxia Jin.
}


\institute{Walid Shalaby \at
              Department of Computer Science \\ 
              University of North Carolina at Charlotte \\
              9201 University City Blvd, Charlotte, NC 28223, USA \\
              \email{wshalaby@uncc.edu}           %  \\
           \and
           Wlodek Zadrozny \at
              Department of Computer Science \\ 
              University of North Carolina at Charlotte \\
              9201 University City Blvd, Charlotte, NC 28223, USA \\
              \email{wzadrozn@uncc.edu}           %  \\
           \and
			Hongxia Jin \at
			Samsung Research America \\ 
			665 Clyde Avenue, Mountain View, CA 94043, USA \\
			\email{hongxia.jin@samsung.com}           %  \\
}

\date{Received: 30 January 2018 / Accepted: 2 August 2018}

\maketitle

\begin{abstract}
	Text representations using neural word embeddings have proven effective in many NLP applications. Recent researches adapt the traditional word embedding models to learn vectors of multiword expressions (concepts/entities). However, these methods are limited to textual knowledge bases (e.g., Wikipedia). In this paper, we propose a novel  and simple technique for integrating the knowledge about concepts from two large scale knowledge bases of different structure (Wikipedia, and Probase) in order to learn concept representations. We adapt the efficient skip-gram model to seamlessly learn from the knowledge in Wikipedia text and Probase concept graph. We evaluate our concept embedding models on two tasks: 1) analogical reasoning, where we achieve a state-of-the-art performance of 91\% on semantic analogies, 2) concept categorization, where we achieve a state-of-the-art performance on two benchmark datasets achieving categorization accuracy of 100\% on one and 98\% on the other. Additionally, we present a case study to evaluate our model on unsupervised argument type identification for neural semantic parsing. We demonstrate the competitive accuracy of our unsupervised method and its ability to better generalize to out of vocabulary entity mentions compared to the tedious and error prone methods which depend on gazetteers and regular expressions.
\keywords{Entity \& Concept Embeddings \and Entity Identification \and Concept Categorization \and Skip-gram \and Probase \and Knowledge Graph Representations
}

\end{abstract}
\begin{figure*}[t]
	\centering
	%\fbox{
	\includegraphics[width=11.5cm,height=10cm,keepaspectratio]{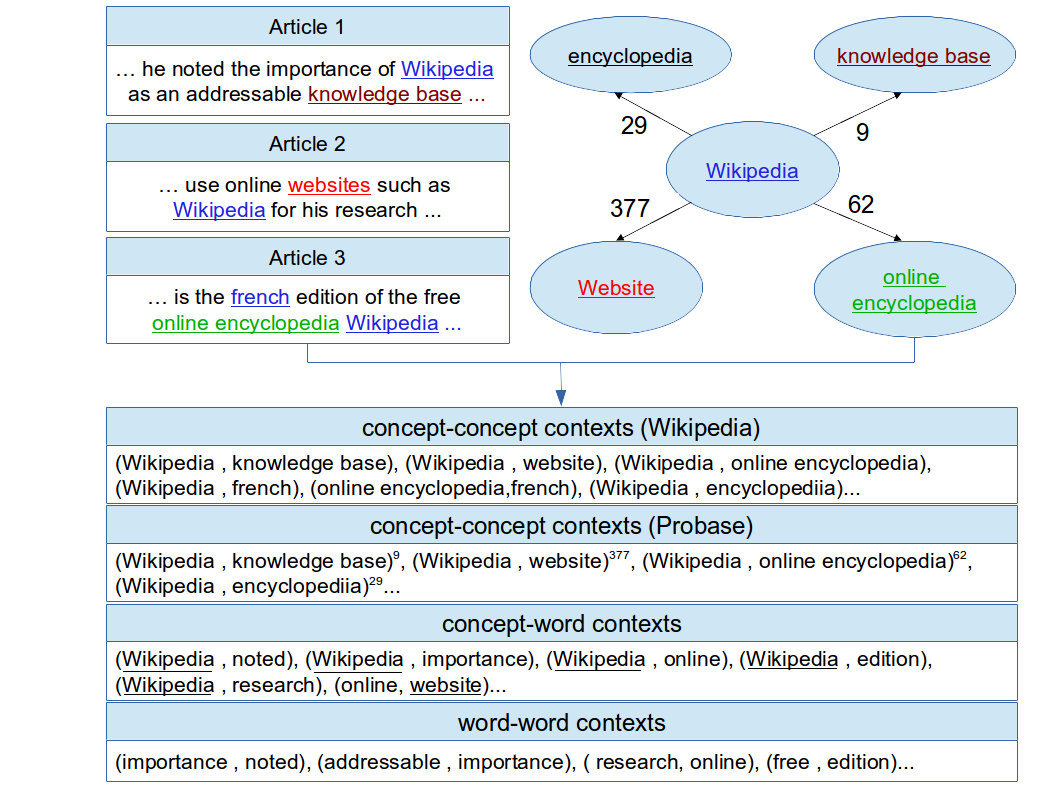}
	%}
	\caption{Integrating knowledge from Wikipedia text (left) and Probase concept graph (right). Local concept-concept, concept-word, and word-word contexts are generated from both KBs and used for training the skip-gram model.}
	\label{examples}
\end{figure*} 

\section{Introduction}
Vector-based semantic representation models are used to represent textual structures (words, phrases and documents) as multidimensional vectors. Typically, these models utilize textual corpora and/or Knowledge Bases (KBs) in order to extract and model real-world knowledge. Once acquired, any given text structure is represented as a real-valued vector in the semantic space. The goal is thus to accurately place semantically similar structures close to each other in that semantic space, while placing dissimilar structures far apart.

Recent neural-based methods for learning word vectors (embeddings) have even succeeded in capturing both syntactic and semantic regularities using simple vector arithmetic (\cite{mikolov2013efficient,mikolov2013distributed,pennington2014glove}). For example, inferring analogical relationships between words: {\it vec(king)-vec(man)+vec(woman)=vec(queen)}. This indicates that the learned vector dimensions encode meaningful multi-clustering for each word.

Word vectors suffer significant limitations. First, each word is assumed to have a single meaning regardless of its context and thus is represented by a single vector in the semantic space (e.g., {\it charlotte (city)} vs. {\it charlotte (given name)}). Second, the space contains vectors of single words only. Vectors of multiword expressions (MWEs) are typically obtained by averaging the vectors of individual words. However, this would often produce inaccurate representations especially if the meaning of the MWE is different from the composition of meanings of its individual words (e.g., {\it vec(north carolina)} vs. {\it vec(north)+vec(carolina)}. Additionally, mentions that are used to refer to the same concept would have different embeddings (e.g., {\it u.s., america, usa}), and the model might not be able to place those individual vectors in the same sub-cluster, especially the rare surface forms.

To address these limitations, a lot of research interest has been focusing on learning distributed representations of concepts and entities which are lexical expressions (single or multiword) that denote an idea, event, or an object and have a set of properties. Typically each concept has an entry in a KB (e.g., an article in Wikipedia or a node in knowledge graph). Such entity embeddings models utilize text KBs (e.g., Wikipedia) or a triple-based KBs (e.g., DBpedia and Freebase) in order to learn entity vectors. Broadly speaking, existing methods can be divided into two categories. First, methods that learn embeddings of KB concepts only (\cite{hu2015entity,zwicklbauer2016robust,li2016joint,ristoski2016rdf2vec}). Second, methods that jointly learn embeddings of words and concepts in the same semantic space (\cite{wang2014knowledge,fang2016entity,yamada2016joint,camacho2016nasari,fang2016entity,cao2017bridge,shalaby2017learning,phan2017neupl}). % Additionally, several other methods were proposed to learn embeddings of entities and relationships from knowledge graphs (see \cite{nguyen2017overview} for detailed overview). 

In this paper, we introduce an effective approach for jointly learning word and concept vectors from two large scale KBs of different modalities: a text KB (Wikipedia) and a graph-based concept KB (Microsoft concept graph\footnote{https://concept.research.microsoft.com} (aka Probase)). We adapt skip-gram, the popular local context window method \cite{mikolov2013distributed}, to integrate the knowledge from both KBs. As shown in Figure \ref{examples}, three key properties differentiate our approach from existing methods. First, we generate word and concept contexts from their raw mentions in the Wikipedia text. This makes our model extensible to other text corpora with annotated concept mentions. Second, we model Probase as a weighted undirected KB graph, exploiting the co-occurrence counts between pairs of concepts. This allows us to generate more concept-concept contexts during training, and subsequently learn better concept vectors for rare and infrequent concepts in Wikipedia. Third, to our knowledge, this work is the first to combine knowledge from two KBs of different modalities (Wikipedia and Probase) into a unified representation. 

We evaluate the generated concept vectors intrinsically on two tasks: 1) analogical reasoning where we achieve a state-of-the-art accuracy of 91\% on semantic analogies, 2) concept categorization on two datasets, where we achieve 100\% accuracy on one dataset and 98\% accuracy on the other. We also present a case study to analyze the impact of using our concept vectors for unsupervised argument type identification with semantic parsing as an end-to-end task. The results show competitive performance of our unsupervised method compared to the tedious and error prone argument type identification methods which depend on gazetteers and regular expressions. The analysis also shows superior generalization performance on utterances containing out of vocabulary (OOV) mentions.

We make our concept vectors and source code publicly available\footnote{https://sites.google.com/site/conceptembeddings/} for the research community for further experimentation and replication.

\section{Learning Concept Embeddings}
\label{learning}
\subsection{Skip-gram}
\label{skipgram}
We learn continuous vectors of words and entities by building upon the skip-gram model of \cite{mikolov2013distributed}. In the conventional skip-gram model, a set of contexts are generated by sliding a context window of predefined size over sentences of a given text corpus. The vector representation of a target word is learned with the objective to maximize the ability of predicting surrounding words of that target word. 

Formally, given a training corpus of $V$ words $w_1, w_2, ..., w_V$. The skip-gram model aims to maximize the average log likelihood probability:
\begin{equation}
\frac{1}{V} \sum_{i=1}^{V}{\sum_{-s \le j \le s,j \ne 0}{\log \ p(w_{i+j}|w_i)}}
\end{equation}

\noindent where $s$ is the context window size, $w_i$ is the target word, and $w_{i+j}$ is a surrounding context word. The softmax function is used to estimate the probability $p(w_{O}|w_I)$ as follows:
\begin{equation}
p(w_{O}|w_I) = \frac{\exp (\mathbf{v}_{w_O}^\intercal \mathbf{u}_{w_I})}{\sum_{w=1}^{V}{\exp(\mathbf{v}_w^\intercal \mathbf{u}_{w_I})}}
\end{equation}

\noindent where $\mathbf{u}_w$ and $\mathbf{v}_w$ are the input and output vectors respectively, and $V$ is the vocabulary size. \citet{mikolov2013distributed} proposed hierarchical softmax and negative sampling as efficient alternatives to approximate the softmax function (which becomes computationally intractable when $V$ becomes huge).

\subsection{Learning from Text}
\label{modeling-text}
Our approach genuinely learns distributed concept representations by generating concept contexts from mentions of those concepts in large encyclopedic text KBs such as Wikipedia. Utilizing such annotated KBs eliminates the need to manually annotate concept mentions and thus comes at no cost.

Here we propose learning the embeddings of both words and concepts jointly. First, all concept mentions are identified in the given corpus. Second, contexts are generated for both words and concepts from other surrounding words and other surrounding concepts as well. After generating all the contexts, we use the skip-gram model to jointly learn embeddings of words and concepts. Formally, given a training corpus of $V$ words $w_1, w_2, ..., w_V$. We iterate over the corpus identifying words and concept mentions and thus generating a sequence of $T$ tokens $t_1, t_2,...t_T$ where $T<V$ (as multiword concepts will be counted as one token). Afterwards we train the a skip-gram model aiming to maximize:
\begin{equation}
\label{eqn3}
\mathscr{L}_t = \frac{1}{T} \sum_{i=1}^{T}{\sum_{-s \le j \le s,j \ne 0}{\log \ p(t_{i+j}|t_i)}}
\end{equation}

\noindent where as in the conventional skip-gram model, $s$ is the context window size. Here, $t_i$ is the target token which would be either a word or a concept mention, and $t_{i+j}$ is a surrounding context word or concept mention. 

\subsection{Learning from Concept Graph}
\label{modeling-graph}
We employ Microsoft concept graph (Probase), a large scale probabilistic KB of millions of concepts and their relationships (basically is-a hierarchy). Probase was created by mining billions of Web pages and search logs of Microsoft's Bing\footnote{https://www.bing.com/} repository using syntactic patterns. The concept KB was then leveraged for text conceptualization to support text understanding tasks such as clustering of Twitter messages and News titles (\cite{song2011short,song2015open}), search query understanding (\cite{wang2015query}), short text segmentation (\cite{hua2015short}), and term similarity (\cite{kim2013context}). 

Probase has a different structure (or modality) than Wikipedia because the knowledge is organized as a graph whose nodes are concepts and edges represent a weighted is-a relationship between pairs of concepts. Formally, we model Probase as a 4-tuple graph $G=(C, E, \mathscr{T}_C, \mathscr{T}_E)$ such that:
\begin{itemize}[topsep=0pt]
	\itemsep0em
	\item $C$ is a set of vertices representing concepts.
	\item $E$ is a set of edges (arcs) connecting pairs of concepts.
	\item $\mathscr{T}_C$ is a finite set of tuples representing global statistics of each concept (i.e. its total occurrences).
	\item $\mathscr{T}_E$ is a finite set of tuples representing co-statistics of each edge connecting pairs of concepts (i.e. their co-occurrence count).
\end{itemize}

Under this representation, location information is lost. Therefore the context of each concept can be defined by the set of its neighbors in the graph. Formally, the skip-gram optimization function would be maximizing:
%\begin{equation}
%\mathscr{L}_p = \frac{1}{|C|} \sum_{i=1}^{|C|}{\sum_{(c_i,c_j)\in E}\sum_{l=1}^{n_{c_i,c_j}}{{\log \ p(c_j|c_i)}}}
%\end{equation}

\begin{equation}
\mathscr{L}_p = \frac{1}{|C|} \sum_{i=1}^{|C|}{\sum_{(c_i,c_j)\in E}{{\log \ p(c_j|c_i)}}}
\end{equation}

\noindent Note that, while maximizing $\mathscr{L}_p$, the number of training examples generated from $(c_i,c_j)\in E$, is equal to their co-occurrence count $n_{c_i,c_j}$. The incorporation of the concept-concept co-occurrence counts in Probase will result in a dynamic adjustment to the overall likelihood $\mathscr{L}_p$ depending on the counts between pairs of concepts. For example, for highly related concepts the co-occurrence count will be high, and so will be their contribution to $\mathscr{L}_p$ and vice versa. Thus Probase provides another source of conceptual knowledge to generate more concept-concept contexts, and subsequently learn better concept representations.

\subsection{Data and Model Training}
\subsubsection{Wikipedia} 
We utilized the Wikipedia dump of August 2016\footnote{http://dumps.wikimedia.org/enwiki/\label{wiki2016}}, which had $\sim$7 million articles. We extracted articles plain text discarding images and tables. We also discarded \textit{References} and \textit{External links} sections (if any). We pruned articles not under the main namespace\footnote{Articles which are prefixed with a string then colon before the title name}. Eventually, our corpus contained $\sim$5 million articles in total. 
We preprocessed each article replacing all its references to other Wikipedia articles with the their corresponding article IDs. In case any of the references is a title of a redirect page, we used the page ID of the original page to ensure that all concept mentions were normalized to their article IDs. 

\subsubsection{Microsoft Concept Graph (Probase)} 
We used Probase data repository\footnote{https://concept.research.microsoft.com/Home/Download} which contained $\sim$5 million unique concepts, $\sim$12 million unique instances, and $\sim$85 million is-a relationships. We followed a simple exact string matching between Wikipedia article titles and Probase concept names in order to align the concepts in both KBs and generate the final concepts set.

\subsubsection{Training}
We call our model Concept Multimodal Embedding (CME). During training, we jointly train our model to maximize $\mathscr{L}  = \mathscr{L}_t + \mathscr{L}_p$, which as mentioned before is estimated using the softmax function. Although it is possible to use weighted sum of $\mathscr{L}_t$ and $\mathscr{L}_p$, we opted using unweighted sum as it is simpler to train, and will not to introduce an extra hyperparameter to the learning model. Thus, we let the model learn the best combination between $\mathscr{L}_t$ and $\mathscr{L}_p$ based on the global words/concepts counts and local co-occurrences between pairs of them.

Following \citet{mikolov2013distributed}, we utilize negative sampling to efficiently approximate the softmax function by replacing every $\log \ p(w_O|w_I)$ term in the softmax function (equation 2) with:
\begin{equation}
\log \sigma(\mathbf{v}_{w_O}^\intercal \mathbf{u}_{w_I}) + \sum_{g=1}^{k}{\E_{w_s\sim P_n(w)} [\log \sigma(-\mathbf{v}_{w_g}^\intercal \mathbf{u}_{w_I})]}
\end{equation}

\noindent where $k$ is the number of negative samples drawn for each term, and $\sigma(x)$ is the sigmoid function ($\frac{1}{1+e^{-x}}$). 

We consider global word and concept statistics when generating the negative samples for training. As in \citet{mikolov2013distributed}, we implement the downsampling trick where words with normalized frequency ($>$10\textsuperscript{-3}) are downsampled. For each training sample, we sample 5 noisy words/concepts as negatives from the uniform distribution raised to 3/4rd power. 

For text learning, we use a context window of size 9. We set the vector size to 500 dimensions and train the model for 10 iterations using 12 cores machine with 64GB of RAM. Our model takes $\sim$15 hours to train. The total vocabulary size is $\sim$12.7 million including words and concepts.

\section{Evaluation}
\subsection{Analogical Reasoning}
\citet{mikolov2013linguistic} introduced this intrinsic evaluation scheme to assess the capacity of the embedding model to learn a vector space with meaningful substructure. Typically, analogies take the form "{\it a to b} is same as {\it c to \_\_?}" where {\it a, b}, and {\it c} are elements of the vocabulary $V$. Using vector arithmetic, this can be answered by identifying {\it d} such that: $d = {arg\ \mbox{max}}_d\ Sim(vec(d),vec(b)-vec(a)+vec(c))$, $\forall d\in V-\{a,b,c\}$, where $Sim$ is a similarity function\footnote{Cosine similarity or dot product if vectors are normalized.}. A good performance on this task indicates the model's ability to learn semantic and syntactic patterns as linear relationships between vectors in the embedding space (\cite{pennington2014glove}).

\subsubsection{Dataset}
We use the word analogies dataset  of \cite{mikolov2013efficient}. The dataset contains 19,544 questions divided into semantic analogies (8,869), and syntactic analogies (10,675). The semantic analogies are questions about country capitals, state cities, country currencies...etc. For example, "{\it cairo} to {\it egypt} is same as {\it paris} to {\it france}". The syntactic analogies are questions about verb tenses, opposites, and adjective forms. For example, "{\it big} to {\it biggest} is same as {\it great} to {\it greatest}". In order to leverage the concept vectors, we first identify the corresponding entity of each analogy word and use its vector. If the word has no corresponding entity or corresponds to a disambiguation page under Wikipedia we use its word vector instead. 

\subsubsection{Compared Systems}
We compare our model to various word and entity embedding methods including:
\begin{enumerate}[topsep=0pt]
	\itemsep0em
	\item \textbf{Word embeddings:} a) Word2Vec$_{sg}$, word embedding model trained on Wikipedia using skip-gram \cite{mikolov2013efficient}, b) Word2Vec$_{sg\_b}$, a baseline model we created by training the skip-gram model on the same Wikipedia dump we used for our CME model, c) GloVe, word embedding model proposed by \cite{pennington2014glove}, and d) GloVe$_b$, same model by \cite{pennington2014glove}, but trained on the same Wikipedia version used by CME without preprocessing, for fair comparison. We use recommended hyperparameter values in \cite{pennington2014glove}.
	\item \textbf{Entity mention embeddings:} MPME, a recent model proposed by \citet{cao2017bridge}. The model jointly learn embeddings of words and entity mentions by training the skip-gram on Wikipedia, and utilizing anchor texts to generate multi-prototype entity mention embeddings.
\end{enumerate}	

\begin{table}[]
	\centering
	{\setlength{\tabcolsep}{0.5em}
		\begin{tabular}{c|cccc} 
			\hline
			\\ [-0.9em]
			\underline{Dataset/Questions}                   & Semantic & Syntactic & All &           \\
			Method & (8,869) & (10,675) & (19,544)          \\ \hline 
			Word2Vec$_{sg}$ & 58  & 61 &  59.5 \\ [0.1em] \hline
			Word2Vec$_{sg\_b}$ & 78.1  & \textbf{62.8} & 69.8  \\ [0.1em] \hline
			Glove & 80.8 & 61.5 & 70.3  \\ [0.1em] \hline
			Glove$_b$ & 69.5 & 32.1 & 49.1  \\ [0.1em] \hline
			MPME  & 71.6  & 54.6 & 63.1  \\ [0.1em] \hline
			CME & \textbf{91.4}  & 61.7 & \textbf{75.2}  \\ [0.1em] \hline								
	\end{tabular}}
	%\egroup
	\caption{Results of analogical reasoning, given
		as percent accuracy (bold indicates best obtained accuracy). Our CME model gives the best result on semantic analogies and higher overall accuracy than all other models.}		
	\label{analogies}
\end{table}

\subsubsection{Results}
We report the accuracy scores of analogical reasoning in Table \ref{analogies}. As we see, our CME model outperforms all other models by significant percentages on the semantic analogies. The closest performing model (Glove) is $\sim$10\% less accurate. Performance on syntactic analogies is still very competitive to Word2Vec$_{sg\_b}$ and GloVe. Overall, our model is $\sim$ 5\% better than the closest performing model.

\subsubsection{Error Analysis} 
%cat /home/wshalaby/work/github/3s/textconceptualization/results/enwiki20160820-words-from-plain-anno-titles.txt.4.4mtitles-10iter-dim500-wind9-cnt1-skipgram1-analogy-eval.log|grep -o -P ", '\w+'\] \w+"|sort|uniq -c > count.txt
%cat /home/wshalaby/work/github/3s/textconceptualization/results/conc-gensim-anno-and-data-concept-titles-10iter-dim500-wind9-cnt1-skipgram1-analogy-eval-conceptualized.log|grep -o -P ", '\w+'\] \w+ \w+"|sort|uniq -c > count-conceptualized.txt
Local context window models like ours generally perform better on semantic analogies than syntactic ones. This indicates that syntactic regularities in most textual corpora are more difficult to capture, using embeddings, than semantic regularities. A possible reason could be the more morphological variations of verbs and adjectives than nouns. Our model training is even more biased toward capturing semantic relationships between concepts by incorporating knowledge from Probase concept graph. This bias caused our model to produce some semantic predictions on the syntactic analogies compared to the Word2Vec$_{sg\_b}$ baseline, returning a semantically related word to the answer. For instance, our model predicted {\it "fast"} rather than {\it "slows"} 9 times compared to 2 times by Word2Vec$_{sg\_b}$. And {\it "large"} rather than {\it "smaller"} 14 times compared to 1 time by Word2Vec$_{sg\_b}$, Another set of errors were predicting the correct word but with wrong ending especially {\it "ing"}. For instance, {\it "implementing"} rather than {\it "implements"} 27 times compared to 19 time by Word2Vec$_{sg\_b}$. We argue that, despite this bias, our CME model still produces very competitive performance compared to other models on syntactic analogies. And more importantly, emphasizing the semantic relatedness between concepts during training contributes to the significant accuracy gains on the semantic analogies.

\begin{algorithm}
	\scriptsize
	\caption{Classification + Bootstrapping}\label{alg1}
	\KwIn{
		$\mathbf{U}\!=\!\{ (l_1,\mathbf{u}_{l_1}),...,(l_n,\mathbf{u}_{l_n})$\}: labels + embeddings \newline
		$\mathbf{D}\!=\!\{ (d_1,\mathbf{v}_{d_1}),...,(d_m,\mathbf{v}_{d_m})$\}: instances + embeddings \newline
		N: number of bootstrap instances
	}
	%\KwOut{how to write algorithm with \LaTeX2e }
	\KwResult{
		$\mathbf{L}\!=\!\{...,(d_i,l_j),...\}$: label assignment for each instance
	}
	%initialization\;
	%\While{condition}{while-block}
	\Repeat{$\mathbf{D} = \phi$ \Comment{no more instances to classify}}{
		$\mathbf{candidates}\gets\{ l_1:\phi,...,l_n:\phi$\}\\
		\ForEach{$(d,\mathbf{v}_{d})\in\mathbf{D}$}{
			$d_{max\_sim} = 0$\\
			$d_{max\_label} = null$\\
			\ForEach{$(l,\mathbf{u}_{l})\in\mathbf{U}$}{
				$sim_{l} = Sim(\mathbf{v}_{d},\mathbf{u}_{l})$\\
				\If{$sim_{l}>d_{max\_sim}$}
				{
					$d_{max\_sim} = sim_{l}$\\
					$d_{max\_lebel} = l$					
				}
			}
			add ($d,d_{max\_sim}$) to $\mathbf{candidates}[l]$\\
		}
		\ForEach{$(l,\mathbf{candidates}_{l})\in\mathbf{candidates}.items$}{
			%\For{$\_\ \ \mbox{in}\ \ 1..N$}{
			\Repeat{N highest scored instances added} {
				$score_{max} = 0$\\
				$d_{max} = null$\\
				\ForEach{$(d,score_{d})\in\mathbf{candidates}_{l}$}{
					\If{$score_{d}>score_{max}$}{
						$score_{max} = score_{d}$\\
						$d_{max} = d$ \Comment{most similar instance so far}
					}
				}
				add $(d_{max},l)$ to $\mathbf{L}$\Comment{assign class label}\\
				$\mathbf{u}_{l} \gets \mathbf{u}_{l}  + \mathbf{v}_{d}$\Comment{bootstrap label embedding}\\
				remove $d$ from $\mathbf{candidates}_{l}$\\
				remove $d$ from $\mathbf{D}$\\
			}
			%$\bar{D}\gets$ N instances with highest scores\;
		}
	}
	
	%\Procedure{Euclid}{$a,b$}\Comment{The g.c.d. of a and b}
\end{algorithm}

%\begin{algorithm}[t]
%	\scriptsize
%	\caption{Classification + Bootstrapping}\label{alg1}
%	\KwIn{
%		$\mathbf{U}\!=\!\{ (l_1,\mathbf{u}_{l_1}),...,(l_n,\mathbf{u}_{l_n})$\}: labels + embeddings \newline
%		$\mathbf{D}\!=\!\{ (d_1,\mathbf{v}_{d_1}),...,(d_m,\mathbf{v}_{d_m})$\}: instances + embeddings \newline
%		N: number of bootstrap instances
%	}
%	%\KwOut{how to write algorithm with \LaTeX2e }
%	\KwResult{
%		$\mathbf{L}\!=\!\{...,(d_i,l_j),...\}$: label assignment for each instance
%	}
%	%initialization\;
%	%\While{condition}{while-block}
%	\Repeat{$\mathbf{D} = \phi$ \Comment{no more instances to classify}}{
%		$\mathbf{candidates}\gets\{ (l_1,\phi),...,(l_n,\phi)$\}\\
%		\ForEach{$(d_i,\mathbf{v}_{d_i})\in\mathbf{D}$}{
%			$l_j,score_{d_i} \gets arg\ \mbox{max}_j\ Sim(\mathbf{v}_{d_i},\mathbf{u}_{l_j})$\\
%			add ($d_i,score_{d_i}$) to $\mathbf{candidates}_{l_j}$
%		}
%		\ForEach{$(l_j,\mathbf{candidates}_{l_j})\in\mathbf{candidates}$}{
%			%\For{$\_\ \ \mbox{in}\ \ 1..N$}{
%			\Repeat{N highest scored instances added} {
%				$d_i \gets \mbox{max}_{score_{d_i}}(\mathbf{candidates}_{l_j})$\\
%				add $(d_i,lj)$ to $\mathbf{L}$\Comment{assign class label}\\
%				$\mathbf{u}_{l_j} \gets \mathbf{u}_{l_j}  + \mathbf{v}_{d_i}$\Comment{bootstrap label embedding}\\
%				remove $d_i$ from $\mathbf{candidates}_{l_j}$\\
%				remove $d_i$ from $\mathbf{D}$\\
%			}
%			%$\bar{D}\gets$ N instances with highest scores\;
%		}
%	}
%	
%	%\Procedure{Euclid}{$a,b$}\Comment{The g.c.d. of a and b}
%\end{algorithm}

\subsection{Concept Learning}
Concept learning is a cognitive process which involves classifying a given concept/entity to one or more candidate categories (e.g., {\it "milk"} as {\it beverage, dairy product, liquid}...etc). This process is also known as {\it concept categorization}\footnote{In this paper, we use concept learning and concept categorization interchangeably} \cite{li2016joint}. 

Automated concept categorization can be viewed through both intrinsic and extrinsic evaluation. Intrinsic because a "good" embedding model would generate clusters of concepts belonging to the same category, and optimally place the category vector at the center of its instances cluster. And extrinsic as the embedding model could be leveraged in many knowledge modeling tasks such as {\it KB construction} (creating new concepts), {\it KB completion} (inferring new relationships between concepts), and {\it KB curation} (removing noisy or assessing weak relationships). 

Similar to \citet{li2016joint}, we assign a given concept to a target category using Rocchio classification (\cite{rocchio1971relevance}), where the centroid of each category is set to the category's corresponding embedding vector. Formally, given a set of $n$ candidate concept categories $G = \{g_1, ..., g_n\}$, an instance concept $c$, an embedding function $f$, and a similarity function $Sim$, then $c$ is assigned to the $i$th category $g_i$ such that $g_i = arg\ \mbox{max}_i\ Sim(f(g_i),f(c))$. Under our CME model, the embedding function $f$ would always map the given concept to its vector.

\subsubsection{Bootstrapping} 
We leverage bootstrapping in order to improve the categorization accuracy without the need for labeled data. In the context of concept learning, we start with the vectors of target category concepts as a prototype view upon which categorization assignments are made (e.g., {\it vec(bird), vec(mammal)...etc}). We leverage bootstrapping by iteratively updating this prototype view with the vectors of concept instances we are most confident. For example, if {\it "deer"} is closest to {\it "mammal"} than any other instance in the dataset, then we update the definition of {\it "mammal"} by performing {\it vec(mammal)+=vec(deer)}, normalize it, and repeat the same operation for other categories as well. This way, we adapt the initial prototype view to better match the specifics of the given data. %Although bootstrapping is a time consuming process, we argue that, using dense vectors for representing concepts makes bootstrapping more appealing. As updating the category vector with an instance vector could be performed through optimized vector arithmetic which is available in most modern machines. 
Algorithm \ref{alg1} presents the pseudocode for performing  concept categorization with bootstrapping. In our implementation, we bootstrap the category vector with vectors of the most similar $\mathbf{N}$ instances at a time. Another implementation option might be defining a threshold and bootstrapping using vectors of $\mathbf{N}$ instances if their similarity scores exceed that threshold. %We set $\mathbf{N}=1$ in our experiments.

\subsubsection{Datasets}
As in \citet{li2016joint}, we utilize two benchmark datasets: 1) Battig test (\cite{baroni2010distributional}), which contains 83 single word concepts (e.g., {\it cat, tuna, spoon..etc}) belonging to 10 categories (e.g., {\it mammal, fish, kitchenware..etc}), and 2) DOTA, which was created by \citet{li2016joint} from Wikipedia article titles (entities) and category names (categories). DOTA contains 300 single-word concepts (DOTA-single) (e.g., {\it coffee, football, semantics..etc}), and (150) multiword concepts (DOTA-mult)  (e.g., {\it masala chai, table tennis, noun phrase..etc}). Both belong to 15 categories (e.g., {\it beverage, sport, linguistics...etc}). Performance is measured in terms of the ability of the system to assign concept instances to their correct categories.

\subsubsection{Compared Systems}
We compare our model to various word, entity and category embedding methods including:
\begin{enumerate}[topsep=0pt]
	\itemsep0em
	\item \textbf{Word embeddings:} \citet{collobert2011natural} model (WE$_{Senna}$) trained on Wikipedia. Here vectors of multiword concepts are obtained by averaging their individual word vectors.
	\item \textbf{MWEs embeddings:} \citet{mikolov2013distributed} model (WE$_{Mikolov}$) trained on Wikipedia. This model jointly learns single and multiword embeddings where MWEs are identified using corpus statistics.
	\item \textbf{Entity-category embeddings:} which include \citet{bordes2013translating} embedding model (TransE). This model utilizes relational data between entities in a KB as triplets in the form (entity, relation, entity) to generate representations of both entities and relationships. \citet{li2016joint} implemented three variants of this model (TransE$_1$, TransE$_2$, TransE$_3$) to generate representations for entities and categories jointly. Two other models introduced by \citet{li2016joint} are CE and HCE. CE generates embeddings for concepts and categories using category information of Wikipedia articles. HCE extends CE by incorporating Wikipedia's category hierarchy while training the model to generate concept and category vectors.
	\itemsep0em
	\item \textbf{Other baselines:} we created three baselines: a) WE$_{b}$, has word embeddings only and was obtained by training the skip-gram model on the same Wikipedia dump we used for our CME model (cf. equation 1), b) Wiki-cc$_{b}$, has concept embeddings only and was obtained by first preprocessing Wikipedia to remove all non-concept tokens, and then training the skip-gram model on concept-concept contexts (cf. equation 3 where each token $t$ is a concept mention), and c) Probase-cc$_{b}$, has concept embeddings only and was obtained by training the adapted skip-gram model on Probase concept graph (cf. equation 4). 
	
	These baselines are meant to quantify and analyze the contribution of each type of information individually. Specifically, entity-entity in Wikipedia conceptual contexts, entity-entity in Probase knowledge graph, and word-word in Wikipedia raw contexts.
	
\end{enumerate}

\begin{table*}[t]
	\centering
	{\setlength{\tabcolsep}{0.5em}
		\begin{tabular}{c|cccc} 
			\hline
			\\ [-0.9em]
			\underline{Dataset/Instances}                   & Battig & DOTA-single & DOTA-mult & DOTA-all          \\
			Method & (83) & (300) & (150) & (450)          \\ \hline 
			WE$_{Senna}$ & 44  & 52 & 32 & 45 \\ [0.1em] \hline
			WE$_{Mikolov}$  & 74  & 72 & 67 & 72 \\ [0.1em] \hline
			TransE\textsubscript{1}  & 66  & 72 & 69 & 71 \\ [0.1em] \hline
			TransE\textsubscript{2}  & 75  & 80 & 77 & 79 \\ [0.1em] \hline
			TransE\textsubscript{3}  & 46  & 55 & 52 & 54 \\ [0.1em] \hline
			CE  & 79  & 89 & 85 & 88 \\ [0.1em] \hline
			HCE  & 87  & 93 & 91 & 92 \\ [0.1em] \hline
			WE$_b$  & 77 & 93 & 86 & 91 \\ [0.1em]
			+bootstrap  & 88 & 97 & 86 & 90 \\ 
			\hline
			Wiki-cc$_b$  & 72 & 90 & 80 & 87 \\ [0.1em]
			+bootstrap  & 81 & 91 & 86 & 87 \\ 
			\hline
			Probase-cc$_b$  & 73 & 65 & 70 & 67 \\ [0.1em]
			+bootstrap  & 95 & 78 & 81 & 83 \\ 
			\hline
			CME  & 94 & 91 & 88 & 90 \\ [0.1em]
			+bootstrap  & \textbf{100} & \textbf{99} & \textbf{95}  & \textbf{98} \\ 
			\hline
			%CRC  & 83  & 91 & 88 & 90 \\ [0.1em]
			%+bootstrap  & \textbf{89}  & \textbf{98} & \textbf{95} & \textbf{97} \\ 
			%\hline
	\end{tabular}}
	%\egroup
	\caption{Results of the concept categorization task, given
		as percent accuracy (bold indicates best obtained accuracy). Our CME model with bootstrapping gives the best results outperforming all other models and baselines.}		
	\label{concept-categorization}
\end{table*}

\subsubsection{Results}
We report the accuracy scores of concept categorization\footnote{From a multi-class classification perspective, the accuracy scores would be equivalent to the clustering purity score as reported in \citet{li2016joint}.} in Table \ref{concept-categorization}. 
Accuracy is calculated by dividing the number of correctly classified concepts by the total number of concepts in the given dataset. Scores of all non-baseline methods are obtained from \citet{li2016joint}. As we can see in Table \ref{concept-categorization}, our CME+bootstrap model outperforms all other models and baselines by significant percentages. It even achieves 100\% accuracy on the Battig dataset. With single word concepts, CME achieves the best performance on Battig and competitive performance to WE$_b$ on DOTA-single. When it comes to multiword concepts, our CME model comes second after HCE. In general, baselines which depend only on pure concept-concept contexts (Wiki-cc$_b$ and Probase-cc$_b$) perform worse than the word-word contexts baseline (WE$_b$). This indicates the significance of the full concept contextual information obtained when including both other nearby words and other nearby concepts while learning target concept representation.

\subsubsection{Analysis} {\it Is bootstrapping a magic bullet?} A first look at the results of CME+bootstrap vs. CME might indicate that if bootstrapping is applied to HCE or WE$_b$ which perform better than CME on some datasets, their performance would still be superior. However, the results of WE$_b$+bootstrap show that the margin of performance gains of bootstrapping is not necessarily proportional to the performance of the model without it. For example, WE$_b$+bootstrap performs worse than CME$_b$+bootstrap on DOTA-single, though WE$_b$ was initially better than CME. This means that bootstrapping other better performing models such as HCE might not be as beneficial as it is to CME. The bottom line here is: the model should learn a semantic space with optimal substructures which cluster instances of the same category together, and keep them far from instances of other categories. This is clearly the case with our CME model which ends up having (near-)optimal category vectors with bootstrapping.

\begin{table*}[t]
	\centering
	\scriptsize
	{\setlength{\tabcolsep}{0.5em}
		\begin{tabular}{c|l|l} 
			\hline
			\\ [-0.9em]
			No                   & Utterance & Logical form  \\ \hline 
			1 & where is \textbf{new orleans}& ( lambda \$0 e ( loc:t \textbf{new\_orleans:ci} \$0 ) )  \\ 
			& where is \textbf{ci0} & ( lambda \$0 e ( loc:t \textbf{ci0} \$0 ) )\\ [0.1em] \hline
			2 & what states border the \textbf{mississippi river}& \begin{tabular}{@{}l@{}}( lambda \$0 e ( and ( state:t \$0 ) \\( next\_to:t \$0 \textbf{mississippi\_river:r} ) ) ) \end{tabular} \\ 
			& how many states border \textbf{ri0}& ( count \$0 ( and ( state:t \$0 ) ( next\_to:t \$0 \textbf{ri0} ) ) )\\ [0.1em] \hline
			3 & list flights from \textbf{philadelphia} to \textbf{san francisco} via \textbf{dallas}& \begin{tabular}{@{}l@{}} ( lambda \$0 e ( and ( flight \$0 ) \\( from \$0 \textbf{philadelphia:ci} ) ( to \$0 \textbf{san\_francisco:ci} ) \\( stop \$0 \textbf{dallas:ci} ) ) ) \end{tabular}         \\ 
			& list flight from \textbf{ci0} to \textbf{ci1} via \textbf{ci2} & \begin{tabular}{@{}l@{}} ( lambda \$0 e ( and ( flight \$0 ) ( from \$0 \textbf{ci0} ) \\( to \$0 \textbf{ci1} ) ( stop \$0 \textbf{ci2} ) ) ) \end{tabular} \\ [0.1em] \hline
			4 & flights from \textbf{jfk} or \textbf{la guardia} to \textbf{cleveland} & \begin{tabular}{@{}l@{}} ( lambda \$0 e ( and ( flight \$0 ) ( or ( from \$0 \textbf{jfk:ap} )\\ ( from \$0 \textbf{lga:ap} ) ) ( to \$0 \textbf{cleveland:ci} ) ) ) \end{tabular} \\ 
			& flight from \textbf{ap0} or \textbf{ap1} to \textbf{ci0} & \begin{tabular}{@{}l@{}}  ( lambda \$0 e ( and ( flight \$0 ) ( or ( from \$0 \textbf{ap0} ) \\( from \$0 \textbf{ap1} ) ) ( to \$0 \textbf{ci0} ) ) ) \end{tabular} \\ [0.1em] \hline
			
	\end{tabular}}
	%\egroup
	\caption{Example utterances and their corresponding logical forms from the geography and flights domains. Left, utterances before and after argument type identification. Right, logical forms before and after argument type identification. City is mapped to {\it ci}, Airport to {\it ap}, and River to {\it ri}.}		
	\label{lang2logic-examples}
\end{table*}

\subsection{Argument Type Identification: A Case Study}
In this section, we present a case study to analyze the impact of using our concept vectors for unsupervised argument type identification with semantic parsing as an end-to-end task. In a nutshell, semantic parsing is concerned with mapping natural language utterances into executable logical forms \cite{wang2015building}. The logical form is subsequently executed on a knowledge base to answer the user question. Table \ref{lang2logic-examples} shows some example utterances and their corresponding logical forms from the geography and flights domains.

\subsubsection{Argument Identification} As we can notice from the examples in Table \ref{lang2logic-examples}, user utterances usually contain mentions of entities of various types (e.g., {\it city}, {\it state}, and {\it airport} names). These mentions are typically parsed as arguments in the resulting logical form. Some of these mentions could be rare or even missing in the training data. As noted by \citet{dong2016language}, this problem reduces the model's capacity to learn reliable parameters for such mentions. 

One possible solution is to preprocess the training data, replacing all entity mentions with their type names (e.g., {\it san francisco} to {\it city}, {\it california} to {\it state}...etc). This step allows the model to see more identical input/output patterns during training, and thus better learn the parameters of such patterns. The model would also generalize better to out of vocabulary mentions because the same preprocessing could be done at test time.

\citet{dong2016language} proposed using gazetteers and regular expressions for argument identification. The authors also demonstrated increased accuracy when employing such approach. However, using regular expressions is error prone as the same utterance could be paraphrased in many different ways. In addition, gazetteers usually have low recall, and will not cover many surface forms of the same entity mention. 

In this paper, we embrace argument type identification in a totally unsupervised fashion. The idea is to build upon the promising performance we achieved in concept categorization and apply the same scheme to map entity mentions to their corresponding type names. Our unsupervised argument type identification is a four step process: 1) we predefine target entity types and retrieve their corresponding vectors from our CME model, 2) we identify entity mentions in user utterances (e.g., {\it mississippi river}), 3) we lookup the mention vector in our CME model, and 4) we compute the similarity between the mention vector and each of the predefined target entity types and choose the most similar type if it exceeds a predefined threshold. This scheme is efficient and doesn't require any manually crafted rules or heuristics. The only needed parameter is the similarity threshold which we fix to 0.5 during experiments.

Note that standard off-the-shelf entity recognition systems could help in identifying the entity mentions but not their type names. In domains like flights, we are interested in non standard types such as {\it airports} and {\it airlines}. It is also important to distinguish between {\it city}, {\it state}, and {\it country} mentions in the geography domain and not classifying all instances of these categories as the standard {\it location} type.

\subsubsection{Datasets}
We analyze our unsupervised scheme on two datasets\footnote{We obtained the raw dataset files by contacting the authors of \citet{dong2016language}}
: 1) GEO which contains a total of 880 utterances about U.S. geography \cite{zettlemoyer2012learning}. The dataset is split into 680 training instances and 200 test instances. Here we target identifying five entity types: {\it city}, {\it state}, {\it river}, {\it mountain}, and {\it country}, and 2) ATIS which contains 5,410 utterances about flight bookings split into 4,480 training instances, 480 development instances, and 450 test instances. Here we target identifying six entity types: {\it city}, {\it state}, {\it airline}, {\it airport}, {\it day name}, and {\it month}.

\subsubsection{Model \& Training} We assess the performance of argument type identification by training \citet{dong2016language} neural semantic parsing model\footnote{https://github.com/donglixp/lang2logic}. The model utilizes sequence-to-sequence learning with neural attention (see \cite{dong2016language} for more details). We use the Seq2Seq variant of the model and do not perform any parameter tuning as our purpose is to analyze the performance before and after argument type identification, and not to get a state-of-the-art performance on these datasets.

\begin{table}[]
	\centering
	{\setlength{\tabcolsep}{0.5em}
		\begin{tabular}{c|ccc} 
			\hline
			\\ [-0.9em]
			Dataset   & GEO & ATIS \\ [0.1em] \hline
			w/o Identification & 68.6 & 73.2 \\ [0.1em] \hline
			w/ Identification & 77.1 & 83.7 \\ [0.1em] \hline
	\end{tabular}}
	%\egroup
	\caption{Results of semantic parsing before and after argument type identification, given
		as percent accuracy. Using CME to identify argument types resulted in improved accuracy on both datasets.}
	\label{lang2logic}
\end{table}

\subsubsection{Results} 
We report the parsing accuracy in Table \ref{lang2logic}. Accuracy is defined as the proportion of the input utterances whose logical form is identical to the gold standard. As we can see, our argument type identification scheme resulted in significant accuracy improvements of $\sim$10\% on both datasets. 

We present this experiment as a case study for the utility of our embedding model in an end-to-end task. We don't claim superiority over other embedding techniques here, rather we show that the application of our embedding space to infer is-a relationships can be extended successfully to other application areas including but not limited to: 1) unsupervised argument type identification, and 2) inferring is-a relationship of other categories ({\it city, state, airline, airport, day name...etc}) than those categories in the concept learning datasets (DOTA and Battig).

\subsubsection{Error Analysis}
Training the Seq2Seq semantic parsing model on preprocessed data is clearly beneficial as the results in Table \ref{lang2logic} show. Without argument identification, the model is prone to the out of vocabulary problem. For example, on GEO we spotted 24 test instances with entities not mentioned in the training data (e.g., {\it new jersey}, {\it chattahoochee river}). The same on ATIS with 23 instances. Another source of errors was due to rare mentions. For example, {\it "portland"} appeared once in GEO training data.

Our scheme demonstrated good ability to capture most entity mentions and map them to their correct type names. However, there was some subtle failure cases. For example, in {\it "what length is the mississippi"}, our scheme mapped {\it "mississippi"} to the {\it state}, while it was mapped to the {\it river} in the gold standard logical form. Another example was mapping {\it "new york"} to the {\it city} in {\it "what is the density of the new york"}, while it was mapped to the {\it state} in the gold standard. 

Overall, the results show competitive performance of our unsupervised method compared to the tedious and error prone argument type identification methods. The analysis also shows superior generalization performance when using unsupervised argument identification with utterances containing out of vocabulary and rare mentions.

%geo
%%model-random-raw-args-stem-500-probase 0.77142857142857
%model-random-raw-args-500-probase 0.78571428571429
%model-org-raw-args-500-probase 0.72857142857143
%model-org-raw-args-stem-500-probase.t7 0.76428571428571
%model-org-raw-stem 0.67857142857143
%model-random-raw 0.65357142857143
%model-builtin-raw 66428571428571
%%model-random-raw-stem 0.68571428571429
%atis
%model-org-raw-args-500-probase 0.82589285714286
%model-org-raw-args-stem-500-probase 0.81919642857143
%%model-random-raw-args-stem-500-probase 0.83705357142857
%model-random-raw 0.75223214285714
%%model-random-raw-stem 0.73214285714286
%model-org-raw-stem 0.75892857142857

\section{Related Work}
\label{related}

Neural embedding models have been proposed to learn distributed representations of concepts and entities. \citet{songunsupervised} proposed using the popular Word2Vec model of \cite{mikolov2013efficient} to obtain the embeddings of each concept by averaging the vectors of the concept's individual words. For example, the embeddings of "\emph{Microsoft Office}" would be obtained by averaging the embeddings of "\emph{Microsoft}" and "\emph{Office}" obtained from the Word2Vec model. Clearly, this scheme fails when the semantics of multiword concepts is different from the compositional meaning of their individual words. %such as \emph{Microsoft Office} 
%is different from the semantics of their individual words. % \emph{Microsoft} and \emph{Office}. 

More robust entity embeddings can be learned from the entity's corresponding article and/or from the structure of the employed KB (e.g., its link graph) as in \cite{hu2015entity,li2016joint,yamada2016joint}; and \cite{shalaby2017learning} who all utilize the skip-gram model, but differ in how they define the context of the target concept. However, all these methods utilize one KB only (Wikipedia) to learn entity representations. Our approach, on the other hand, learns better entity representations by exploiting the conceptual knowledge in a weighted KB graph (Probase) and not only from Wikipedia.

Unlike \citet{hu2015entity} and \citet{li2016joint} who learn entity embeddings only, our proposed CME model maps both words and concepts into the same semantic space. 
%Therefore we can easily measure word-word, word-concept, and concept-concept semantic similarities. 
In addition, compared to \citet{yamada2016joint} model which also learns words and entity embeddings jointly, we better model the local contextual information of entities and words in Wikipedia viewed as a textual KB. During training, we generate word-word, word-concept, concept-word, and concept-concept contexts (cf. equation \ref{eqn3}). In \citet{yamada2016joint} model, concept-concept contexts are generated from Wikipedia link graph, and not from their raw mentions in Wikipedia text. 

Exploiting all concept tokens surrounding a target concept allows us, given another corpus with annotated concept mentions, to easily harness concept-concept contexts even if the corpus has no link structure (e.g., news stories, scientific publications, medical guidelines...etc).

Our model is computationally less costly than those of \citet{hu2015entity} and \citet{yamada2016joint} as it requires a few hours rather than days to train using similar computing resources.

Although the learning of the embeddings might seem straightforward, as it uses the standard skip-gram model, we see this as an advantage. On one hand, it allows our training to scale  efficiently to huge vocabulary of words and concepts without the need for a lot of preprocessing (e.g., removing low frequent words and phrases as in \citet{wang2014knowledge,fang2016entity}). On the other hand, to learn from the knowledge graph contexts, we propose simple adaption to the skip-gram model (cf. equation 4), which allows us to use the same dot product scoring function when optimizing for both $\mathscr{L}_t$ and $\mathscr{L}_p$. This is a simpler and more computationally efficient function than the scoring function proposed by previous approaches which learn from knowledge graphs (cf. \cite{fang2016entity}'s equation 1).

\section{Conclusion \& Discussion}
Concepts are lexical expressions (single or multiword) that denote an idea, event or an object and typically have a set of properties associated with it. In this paper, we introduced a neural-based approach for learning embeddings of explicit concepts using the skip-gram model. %Our approach does not require training on a hierarchical concept category graph. 
Our approach learns concept representations from mentions in free text corpora with annotated concept mentions. These mentions even if not available could be obtained through state-of-the-art entity linking systems. We also proposed an effective and seamless addition to the skip-gram learning scheme to learn concept vectors from two large scale knowledge bases of different modalities (Wikipedia, and Probase).

We evaluated of the learned concept embeddings intrinsically and extrinsically. Our performance on the analogical reasoning task produced a new state-of-the-art performance of 91\% on semantic analogies. 

Empirical results on two datasets for performing concept categorization show superior performance of our approach over other word and entity embedding models. 

We also presented a case study to analyze the feasibility of using the learned vectors for argument identification with neural semantic parsing. The analysis shows significant performance gains using our unsupervised argument type identification scheme and better handling of out of vocabulary entity mentions.

To our knowledge, this work is the first to combine knowledge from both Wikipedia and Probase into a unified representation. Our concept space contains all Wikipedia article titles ($\sim$5 million). We use Probase as another source of conceptual knowledge to generate more concept-concept contexts, and subsequently learn better concept vectors. In this spirit, we first filter Probase graph keeping only edges whose both vertices are Wikipedia concepts. Using string matching, $\sim$1 million unique Probase concepts were mapped to Wikipedia articles. Note that we still use the contexts generated from the 5 million Wikipedia concepts, and add to them contexts obtained from the filtered Probase graph. Out of the $\sim$12.7 million vectors in our model, we have $\sim$5 million concept vectors and $\sim$7.7 million word vectors.

One important future improvement is to better match entities from both Wikipedia and Probase. For example, using string edits to increase recall or graph matching techniques to increase precision. Despite using a simple string matching, the performance of our method is superior to other methods utilizing Wikipedia only. It is expected that string matching might produce incorrect mappings. However, it is important to mention that our string matching exploits the redirect pages titles as well as the canonical titles of Wikipedia articles. This increases the recall. For example, in Probase, {\it nyc}, {\it city of new york}, {\it new york city} are all matched with same Wikipedia article {\it New York City}. 

Our initial qualitative analysis shows that it is common to match single-sense Wikipedia concepts ({\it ss-Wiki}) with multi-sense Probase concepts ({\it ms-Pro}). However, in many of these cases, the {\it ms-Pro} is dominated by the {\it ss-Wiki}. For example, the Wikipedia page for {\it Tiger} describes the animal. In Probase, {\it Tiger} is-a {\it Animal} and {\it Tiger} is-a {\it Big cat} has more co-occurrences (917 \& 315 respectively) compared to {\it Tiger} is-a {\it Dance} (1 co-occurrence). Same for {\it Rose} which is described in Wikipedia as flowering plant. In Probase, {\it Rose} is-a {\it Flower} has (906) and {\it Rose} is-a {\it Plant} has (487) co-occurrences compared to {\it Rose} is-a {\it Garden} (10) and {\it Rose} is-a {\it Odor} (5) co-occurrences. We believe this would help generating more consistent contexts from Wikipedia and Probase. On the other hand, such multiple sense concepts in Probase could be leveraged for tasks like sense disambiguation and multi-prototype embeddings, along the lines of \citet{camacho2016nasari}, \citet{iacobacci2015sensembed}, and \citet{mancini2016embedding}. %which is different from the research problem addressed in this work and we leave it for future exploration.

One important aspect of our CME model is its ability to better represent the long tail entities with few mentions. Existing approaches that utilize Wikipedia's link graph treat Wikipedia as unweighted directed KB graph. During training, a context is generated for entities $e_1$ and $e_2$ if $e_1$ has incoming/outgoing link from/to $e_2$. This mechanism poorly represents rare/infrequent Wikipedia concepts which have few incoming links (i.e. few mentions). We, alternatively, exploit Probase link structure modeling it as a weighted undirected KB graph. We also utilize the co-occurrence counts between pairs of concepts (cf. Figure \ref{examples}). Therefore, we generate more concept-concept contexts, resulting in better representations of the long-tail concepts. Consider for example {\it Nightstand} which has in Wikipedia 17 incoming links. In Probase, {\it Nightstand} is-a {\it Furniture}, is-a {\it Casegoods}, and is-a {\it Bedroom furniture} with co-occurrences 47, 47, and 32 respectively. This is a 100+ more contexts than we can generate from Wikipedia. Even for frequent Wikipedia concepts, by exploiting the co-occurrence counts, our model will reinforce concept-concept relatedness from the many contexts obtained from Probase.

Our aim in this work was to combine the knowledge from both Wikipedia and Probase in a seamless and simple way which is scalable (computationally cheap) and effective. The integration learning scheme and the results show that we can achieve these two goals with high degree of success. It principle, it is possible to perform such integration between Wikipedia and Probase contexts in other ways, which may for example distinguish between syntactic and semantic information in these contexts. However, such approaches will require extra preprocessing in order to prepare such contexts. For instance, \cite{levy2014dependency} explored learning word embeddings from contexts generated from a dependency parser. We still claim an advantage over such approaches, because they require costly preprocessing in terms of scalability and effectiveness. As demonstrated by the results, our CME model advances the state-of-the-art on both the analogical reasoning and the concept learning tasks, without the need to do expensive preprocessing or training to learn concept representations.

\begin{acknowledgements}
This work was partially supported by the National Science Foundation under grant number 1624035. Any opinions, findings, and conclusions or recommendations expressed in this material are those of the authors and do not necessarily reflect the views of the National Science Foundation. The authors would like to thank Avik Ray and Yilin Shen from Samsung Research America for their constructive feedback and discussions while developing the case study on the argument type identification task. The authors also appreciate the reviewers valuable and profound comments.
\end{acknowledgements}

%\bibliographystyle{ieeetr}      % basic style, author-year citations
\bibliographystyle{spbasic}      % basic style, author-year citations
\bibliography{concept-embed}   % name your BibTeX data base

%\begin{thebibliography}{}

%\end{thebibliography}

\end{document}